\documentclass[10pt,twocolumn,letterpaper]{article}

\usepackage[pagenumbers]{wacv/wacv} 

\usepackage{graphicx}
\usepackage{amsmath}
\usepackage{amssymb}
\usepackage{booktabs}

\usepackage[pagebackref,breaklinks,colorlinks]{hyperref}

\usepackage[capitalize]{cleveref}
\crefname{section}{Sec.}{Secs.}
\Crefname{section}{Section}{Sections}
\Crefname{table}{Table}{Tables}
\crefname{table}{Tab.}{Tabs.}

\def\wacvPaperID{643} 
\def\confName{WACV}
\def\confYear{2025}

\usepackage{wacv/wacv}
\usepackage{times}
\usepackage{epsfig}
\usepackage{graphicx}
\usepackage{amsmath}
\usepackage{amssymb}
\usepackage{booktabs}

\def\wacvPaperID{643} 

\usepackage{bookmark}
\usepackage{graphicx}
\usepackage{tikz}
\usetikzlibrary{positioning,calc,patterns,snakes}
\usepackage{booktabs}
\usepackage{amsmath}
\usepackage{algorithm}
\usepackage{algpseudocode}
\usepackage{multirow}
\usepackage{ amssymb }
\usepackage[normalem]{ulem}

\usepackage[accsupp]{axessibility}  

\usepackage{hyperref}

\usepackage{orcidlink}

\newcommand{\XT}{\mathbf{X}_T}
\newcommand{\Xt}{\mathbf{X}_t}

\newcommand{\Xtm}{\mathbf{X}_{t-1}}

\newcommand{\Xhatt}{\hat{\mathbf{X}}_t}

\newcommand{\Xhatzero}{\hat{\mathbf{X}}_0}
\newcommand{\Xzero}{\mathbf{X}_0}
\newcommand{\XzeroT}{\mathbf{X}_{0:T}}
\newcommand{\Ut}{\mathbf{U}_t}
\newcommand{\varepsilont}{\boldsymbol{\varepsilon}_t}

\newcommand{\varepsilontXhatt}{\boldsymbol{\varepsilon}_\theta\left(\Xhatt^i, t\right)}

\newcommand{\Var}{\mathrm{Var}}
\newcommand{\setblue}{\color{black}}
\newcommand{\setblack}{\color{black}}
\newcommand{\blue}[1]{\textcolor{black}{#1}}

\newcommand{\tauOneT}{\boldsymbol{\tau}_{1:T}}
\newcommand{\Nstandard}{\mathcal{N}(\mathbf{0}, \mathbf{I})}

\begin{document}

\title{Diffusion Model Guided Sampling with Pixel-Wise Aleatoric Uncertainty Estimation } 
\thispagestyle{empty}

\author{Michele De Vita\\
Friedrich-Alexander-Universität \\
Erlangen-Nürnberg\\
{\tt\small michele.de.vita@fau.de}
\and
Vasileios Belagiannis \\
Friedrich-Alexander-Universität\\
Erlangen-Nürnberg\\
{\tt\small vasileios.belagiannis@fau.de}
}

\maketitle

\begin{abstract}

    Despite the remarkable progress in generative modelling, current diffusion models lack a quantitative approach to assess image quality. To address this limitation, we propose to estimate the pixel-wise aleatoric uncertainty during the sampling phase of diffusion models and utilise the uncertainty to improve the sample generation quality. The uncertainty is computed as the variance of the denoising scores with a perturbation scheme that is specifically designed for diffusion models. We then show that the aleatoric uncertainty estimates are related to the second-order derivative of the diffusion noise distribution. We evaluate our uncertainty estimation algorithm and the uncertainty-guided sampling on the ImageNet and CIFAR-10 datasets. In our comparisons with the related work, we demonstrate promising results in filtering out low quality samples. Furthermore, we show that our guided approach leads to better sample generation in terms of FID scores. 

\end{abstract}

\section{Introduction}
\label{sec:intro}

Recently, diffusion models have made significant progress in producing synthetic images that appear realistic \cite{dhariwal2021diffusion,song2020denoising,ho2020denoising}.
However, the quality of the generated images is not always consistent, and the models may produce artefacts or low-quality samples. Therefore, understanding and quantifying the uncertainty associated with the generated samples is crucial for ensuring the quality of the data, especially in safety-critical applications such as medical imaging \cite{hemsley2020deep, chen2018deep} or autonomous driving \cite{neumeier2024reliable, filos2020can}.

\begin{figure*}[t]
    \centering
    \includegraphics[width=0.9\textwidth]{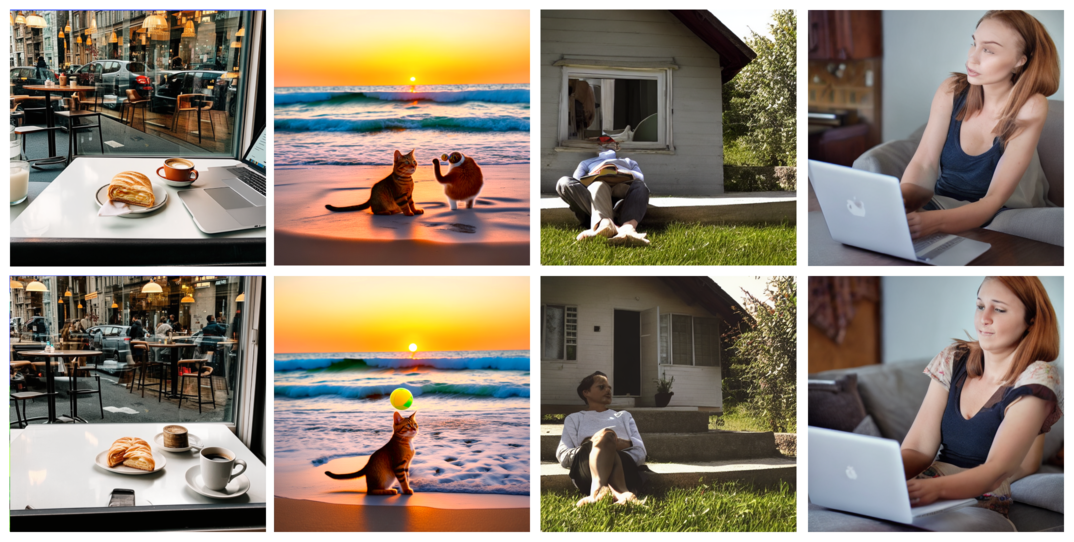}
    \caption{Visual Results I. We provide qualitative samples of uncertainty guidance applied to Stable Diffusion 3 \cite{esser2024scaling} (first two columns) and 1.5 \cite{rombach2022high} (last two columns). In the upper row we present images produced without the uncertainty guidance while the bottom row features images generated with the uncertainty guidance. We can observe that the images with uncertainty guidance present fewer artefacts and more faithful generation}
    \label{fig:sd_comparison}
\end{figure*}

While for established generative models, such as Generative Adversarial Networks (GANs) \cite{goodfellow2020generative} and Variational auto-encoders (VAEs) \cite{kingma2013auto}, there are already a few approaches to obtain uncertainty estimates \cite{saatci2017bayesian, notin2021improving, pmlr-v37-blundell15}, diffusion models remain mostly unexplored. Although it is possible to rely on common uncertainty estimation methods, such as Monte Carlo dropout \cite{gal2016dropout} or ensemble methods \cite{lakshminarayanan2017simple}, these approaches are computationally expensive and not easily applicable to diffusion models. For instance, MC Dropout requires a diffusion model to be trained with dropout, which is quite uncommon and sampling needs to be performed several times. Furthermore, ensemble methods require multiple models to be trained and it is pretty expensive in terms of time budget and computational resources. 
The only method to estimate pixel-wise predictive uncertainty for diffusion models is the recently proposed BayesDiff \cite{kou2023bayesdiff}. Building on the limitations of the aforementioned uncertainty estimation methods, BayesDiff provides an efficient ad-hoc formulation to estimate uncertainty for image generations based on the Last Layer Laplace Approximation (LLLA) \cite{laplace2021}. However, BayesDiff still requires a significant amount of Number of Function Evaluations (NFEs) and does not leverage uncertainty to steer the sampling process. Unlike BayesDiff, we present an approach that is not only computationally more efficient, but more importantly makes use of the uncertainty to guide the generation process towards regions of better sample quality, as illustrated in Fig.~\ref{fig:sd_comparison}.

We propose a training-free and computationally efficient approach to estimate the aleatoric pixel-wise uncertainty during the sampling phase of diffusion models. Our method\blue{\footnote{Our code is available at \url{https://github.com/Michedev/diffusion-uncertainty}.}} estimates the uncertainty as the sensitivity \cite{mi2022training} of multiple data points with the same denoising process. Then, we theoretically show that the proposed uncertainty measure is connected to the second derivative of the noising distribution, providing a solid understanding for our approach. 
Given our uncertainty estimates, pixels with high second-order derivatives are more susceptible to changes during sampling, representing features or details that are more challenging for the model to reconstruct consistently. By directing the sampling process towards high-uncertainty regions, we achieve superior image quality from the same initial conditions. Note that our approach is designed to measure data uncertainty, and thus provides aleatoric pixel-wise uncertainty estimates.

We show the effectiveness of our approach by filtering out low-quality samples in ImageNet \cite{deng2009imagenet} and CIFAR-10 \cite{Krizhevsky09learningmultipleCifar10} datasets. Our approach outperforms existing uncertainty estimation methods in terms of both sample quality and function evaluations on ImageNet. In addition, we demonstrate the generalisation capabilities of our approach by evaluating it on different samplers and neural network architectures. Overall, our contributions are summarised as:\begin{itemize}
    \item We propose a training-free, pixel-wise uncertainty estimation approach for diffusion models. During each sampling step, our algorithm estimates the uncertainty as the variance of multiple generated samples with the same denoising process. 
    \item We show that the uncertainty estimates gives second-order information about the noising distribution. Given this fact, we present an algorithm to guide the sampling phase based on the per-pixel uncertainty estimates.

    \item Our experiments demonstrate state-of-the-art performance compared to previous work on ImageNet and CIFAR-10.  Also, we show that our method improves the quality of generated samples by guiding the diffusion model to focus on areas with low uncertainty.

\end{itemize}

\section{Related Work}

We discuss below the related work on uncertainty estimation, focusing on generative and diffusion models. 

\subsection{Traditional Uncertainty Estimation Methods}

Variational Bayesian Neural Networks (BNNs) \cite{pmlr-v37-blundell15} have been developed to approximate posterior distributions over weights by providing better-calibrated uncertainties and improving the model generalisation, as shown by Wilson \etal \cite{wilson2020bayesian}. However, BNNs can be difficult to train compared to standard neural networks due to optimisation challenges and computational cost. For these reasons, recent approaches have aimed to approximate BNNs more efficiently \blue{\cite{hornauer2022gradient, morales2020activation,pmlr-v80-teye18a,hornauer2023out,wiederer2023joint}}. For instance, Morales-Álvarez \etal \cite{morales2020activation} proposed modelling uncertainty in neural networks by using Gaussian process priors on the activation functions rather than on the weights. Teye \etal \cite{pmlr-v80-teye18a} approximate the uncertainty efficiently using Batch Normalisation \cite{ioffe2015batch}, which is equivalent to approximate inference in Bayesian models.

Another uncertainty estimation method is Monte Carlo dropout (MC-Dropout), proposed by Gal et al. (2016) \cite{gal2016dropout}, which leverages dropout at test time to obtain an approximation of a Bayesian neural network. However, MC-Dropout requires a model trained with dropout and multiple forward passes at test time.
Deep ensembles, proposed by Lakshminarayanan et al. (2017), \cite{lakshminarayanan2017simple} provide a simpler approach by training an ensemble of neural networks with different random initialisations. At test time, the predictions are averaged to obtain the ensemble prediction and variance for uncertainty estimates. Deep ensembles have a higher computational cost due to the training of multiple models, but are easier to optimise compared to BNNs. 
Snapshot Ensembles, proposed by Huang \etal \cite{huang2017snapshot}, is a method to train an ensemble of neural network models at no additional cost compared to training a single model. The approach relies on the ability of the optimisers to escape local minima using a cyclic learning rate to save several snapshots of the models.

Although the above approaches to uncertainty estimation can be applied to any type of parametric model, they are either computationally expensive or with strict requirements on the model architecture and, therefore not easily applicable to diffusion models.

\subsection{Uncertainty Estimation for Generative Models}

Recent approaches explore uncertainty estimation to identify low quality and out-of-distribution samples from generative models \cite{hornauer2023heatmap,hornauer2023out}. BayesGAN, by Saatci \etal \cite{saatci2017bayesian}, incorporates uncertainty estimation into generative adversarial networks (GANs) \cite{goodfellow2020generative} by placing posterior distributions over the generator and discriminator parameters. However, the computational overhead of posterior sampling with stochastic gradient Hamiltonian Monte Carlo limits its scalability and it does not provide pixel-wise estimates.

Grover \etal \cite{pmlr-v89-grover19a} proposes Uncertainty auto-encoders, an auto-encoder based approach that is trained to maximise the mutual information between the input and the latent representation. Similarly, recent work \cite{sagar2022uncertainty} utilises auto-encoders to segment tumor regions in medical images, while quantifying the uncertainty of the segmentation. 
A special type of auto-encoders are Variational auto-encoders (VAEs), by Kingma \etal (2013) \cite{kingma2013auto}. Unlike regular auto-encoders, VAEs are inherently stochastic as their latent space encodes a distribution rather than a fixed value. By sampling multiple times from the latent space, VAEs can provide pixel-wise uncertainty estimation of the data. Notin \etal (2021) \cite{notin2021improving} rely on the uncertainty estimates from the VAEs to filter out low-quality samples from the generations.  While VAEs by An \etal \cite{An2015VariationalAB} provide basic uncertainty information by optimising the reconstruction probability, diffusion models are more powerful in terms of log-likelihood approximation, and consequently there is more interest in developing uncertainty estimation methods for diffusion models.

However, one critical issue with the diffusion models is their inherent inability to estimate the pixel-wise uncertainty of the generated images. The only approach that measures uncertainty for the diffusion model is BayesDiff \cite{kou2023bayesdiff} proposed by Bao \etal. The paper proposes Last-Layer Laplace Approximation (LLLA) for efficient Bayesian inference of pre-trained score models. It enables the simultaneous generation of images along with pixel-wise uncertainty estimation. However, BayesDiff can estimate the uncertainty only for the generated images, which prohibits the guidance of the generation process. 
Unlike other methods, our method provides uncertainty estimation not only for the generated image, but also during the generation process allowing to guide the generation process.

\section{Method}

We propose an uncertainty estimation approach for the sampling phase of diffusion models, focusing on images $X \in \mathbb{R}^{W \times H \times 3}$, although our method is data-agnostic.

 We then rely on the pixel-wise uncertainty estimate maps to guide the diffusion sampling process. In the following, we present the problem formulation (Sec.~\ref{Sec:Problem}), diffusion models background (Sec.~\ref{sec:diffusion-background}), a discussion of sensitivity (Sec \ref{sec:sensitivity}) and then introduce our uncertainty estimation algorithm (Sec.~\ref{sec:method}) and its connection to the curvature of the noising distribution (Sec.~\ref{Sec:Fischer}). Finally, we make use of the uncertainty to guide the diffusion sampling (Sec.~\ref{sec:improving-generation}).

\subsection{Problem Formulation}\label{Sec:Problem}

Let $\XT \in \mathbb{R}^{W\times H \times 3}$ be sampled from a standard Gaussian distribution. Then the diffusion sampling process iteratively removes the noise $T$ times to produce the image $\Xzero \in \mathbb{R}^{W\times H \times 3}$. While the true posterior distribution  $p_\theta(\varepsilont | \Xt, t)$ is intractable, following \cite{mi2022training} we estimate the uncertainty map $\Ut \in \mathbb{R}^{W\times H \times 3}$ for each sampling step $t$ with $t=\{T, \dots, 0\}$ using sensitivity as an approximation of the posterior variance. Based on the uncertainty map $\Ut$, our goal is to (1) adjust the diffusion model sampling by understanding, which parts of the image are generated at any time step $t$ and (2) utilise the total uncertainty to measure the image quality. Finally, we aim to estimate the pixel-wise diffusion uncertainty map $\Ut$ for each diffusion sampling step without interfering with the training or sampling algorithms of the diffusion model, \ie with a scheduler-agnostic approach.

\subsection{Diffusion Models}
\label{sec:diffusion-background}

Diffusion models learn to generate the data distribution (e.g. images, time-series, latent space etc.) \blue{\cite{dhariwal2021diffusion,ho2020denoising, asthana2024multiconditioned,kingma2021variational}} with a noising process, by gradually adding Gaussian noise to the initial data sample $\Xzero$ according to a predefined variance schedule $\beta_1, ..., \beta_T$. The model is then trained to reverse the noising process~\cite{nielsen2024diffenc,dhariwal2021diffusion,kingma2021variational}, also known as the denoising process. 
For a large value of T, the total number of noising steps, $\XT$ is approximately distributed as a standard Gaussian distribution $\mathcal{N}(\mathbf{0}, \mathbf{I})$. 

\paragraph{Noising}
\label{par:noising}
A single noising step is defined as follows:
\begin{equation}
    \label{eq:noising}
    q(\Xt | \Xtm) = \mathcal{N}(\Xtm; \sqrt{1-\beta_{t}} \Xt, \beta_{t} \mathbf{I}),
\end{equation}
where $q$ is the noising distribution, $t \in 1 \dots T$ indexes the diffusion steps and $\beta_t \in [0,1]$ is the noise schedule. During the noising process, as $t$ increases, $\Xt$ deviates from the original data distribution towards the standard Gaussian distribution $\mathcal{N}(\mathbf{0}, \mathbf{I})$. The parameter $\beta_t$ controls the variance of the noise added at each step. From Eq.~\ref{eq:noising}, we derive that it is possible to reach $\Xt$ from $ \Xzero $ for any $ t = 1 \dots T $ by reformulating it as:

\begin{equation}
    \label{eq:recursive_noising}
    q(\Xt|\Xzero) = \mathcal{N}(\Xt; \sqrt{\bar{\alpha}_t} \Xzero, (1 - \bar{\alpha}_t) \mathbf{I}),
\end{equation}
where $ \bar{\alpha}_t = \prod_{s=1}^t (1 - \beta_s)$ and $\beta_s$ is the diffusion noise schedule at time-step $t$. To sample from this distribution we utilise the reparametrisation trick \cite{kingma2013auto} as $
    \Xt = \sqrt{\bar{\alpha}_t} \Xzero + (1 - \bar{\alpha}_t) \mathbf{\epsilon}
    $ where $\epsilon\sim\mathbf{\mathcal{N}(0,I)}$

\paragraph{Denoising}

In denoising, the goal is to recover the original data $\Xzero$ from corrupted data $\XT$ by reversing a diffusion process that gradually adds noise. Specifically, DDPMs train a neural network model $\varepsilon_\theta$ with parameters $\theta$ to learn the reverse process of removing noise. 
The single denoising step, that goes from $ \Xt $ to $ \Xtm $ for any $ t = T\dots 1  $ is defined as follows:

\begin{equation}
    \label{eq:denoising}
p_{\theta}(\Xtm | \Xt) = \mathcal{N}(\Xtm; \mu_{\theta}(\Xt, t), \beta_t \mathbf{I}),
\end{equation}

where $\mu_\theta$, the mean of the distribution is given by: 

\begin{equation}
    \label{eq:denoising_mean}
\mu_{\theta}(\Xt, t) = \frac{1}{\sqrt{\alpha_t}} \left( \Xt - \dfrac{\beta_t}{\sqrt{1 - \bar{\alpha}_t}} \varepsilon_\theta(\Xt, t) \right),
\end{equation}
where  $ \bar{\alpha}_t = \prod_{s=1}^t (1 - \beta_s)$ and $\alpha_t = (1 - \beta_t)$ and $\beta_t$ is the diffusion noise schedule at time-step $t$. The denoising score at step $t$ that is computed by the neural network $\varepsilon_\theta$ with parameters $\theta$ is defined as the score term $ \varepsilon_\theta(\Xt, t) $ . 

\paragraph{Score Matching and SDE}

Additionally, the score term $\varepsilon_\theta(\Xt, t)$ is proportional to the gradient of the probability distribution $\nabla_{\Xt} \log q_{\theta} (\Xt | \Xzero )$ as diffusion models resembles a reverse Stochastic Differential Equation \cite{anderson_reverse-time_1982}:
\begin{equation}
\label{eq:reverse_sde}
d\Xt = \left[ -0.5 f(\Xt, t) - g(t)^2 \nabla_{\Xt} \log q(\Xt) \right] dt + g(t) d\bar{w}
\end{equation}
where $f(x, t)$ is the drift coefficient, $g(t)$ is the diffusion coefficient, $q_t(\Xt) = \int p_{\mathcal{D}}(\Xzero) q(\Xt | \Xzero) d\Xzero$  and $\bar{w}$ is the Wiener process. During training, the neural network is optimised to match the score $\nabla_{\Xt} \log q(\Xt|\Xzero) = - \dfrac{\epsilon}{\sigma_t}$ \cite{song2019generative,song2020denoising,song2021scorebased}, where $\epsilon \sim \mathcal{N}(\mathbf{0, I})$ is the aleatoric part of $q(\Xt|\Xzero)$ (see under Eq. \ref{eq:recursive_noising}) and $\sigma_t$ is the noise applied to timestep t. We utilise this match to find the relationship between our uncertainty estimates and the curvature in Section \ref{sec:curvature-log-likelihood-fisher-information}.

\subsubsection{Sampling}
\label{subsubsec:diffusion-sampling}

By sampling from the prior distribution $ \XT \sim N(\mathbf{0}, \mathbf{I}) $ and then iteratively removing the noise $ T $ times using the denoising Eq.~\ref{eq:denoising}, we turn pure Gaussian noise into a new sample $ \Xzero $ that follows the true data distribution. The sampling process is described by the following distributions:
\begin{equation}
    \label{eq:sampling-distribution}
    \begin{aligned}
        p_\theta (\Xzero) & =  \int p_\theta(\Xzero, \mathbf{X}_1, \dots \XT) d\mathbf{x}_{1:T} \\
        & = \int p_\theta(\XzeroT) d\mathbf{x}_{1:T}, \\ 
        &  p_\theta(\XzeroT) = p(\XT) \prod_{t=T}^{1} p_{\theta}(\Xtm | \Xt),
    \end{aligned}
\end{equation}
where $ p(\XT) ~ \sim ~ N(\mathbf{0}, \mathbf{I}) $ and $ p_{\theta}(\Xtm | \Xt) $ is the denoising distribution defined in Eq.~\ref{eq:denoising}.

\begin{figure}
    \centering
    \includegraphics[width=0.48\textwidth]{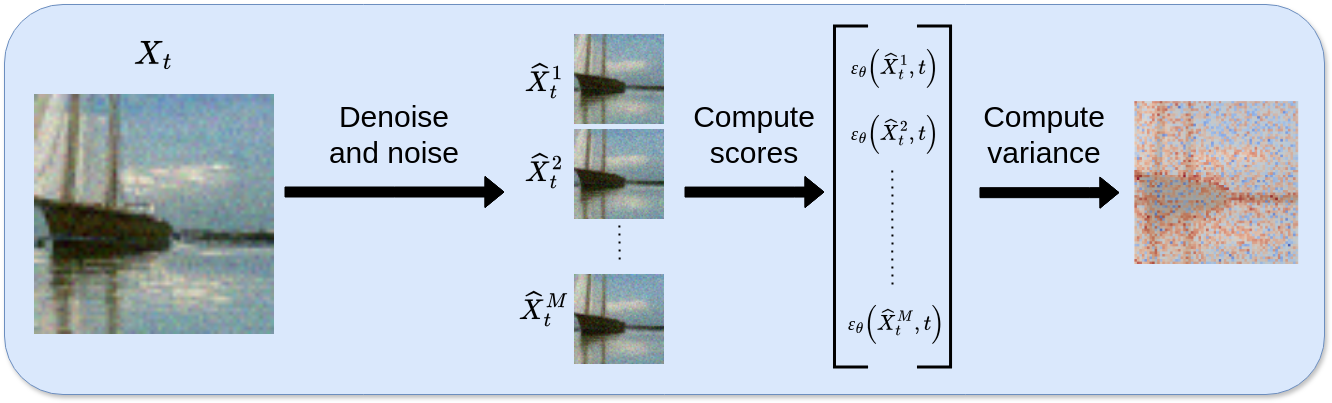}

  \caption{Illustration of our uncertainty estimation algorithm for the timestep \textit{t}. We compute the uncertainty of the denoising process at step $t$ by first computing an approximation of the denoised image $\Xhatzero$ and then sampling from the distribution $q(\Xhatt|\Xhatzero)$ multiple times. The variance of the scores $ \varepsilon_\theta(\Xhatt, t) $ is then computed as the uncertainty of the image at step $t$.}
  \label{fig:infer-noise-diff}

\end{figure}

\subsection{Sensitivity and Uncertainty}
\label{sec:sensitivity}
The proposed uncertainty estimation approach applies sensitivity analysis in the context of diffusion models. Based on the findings of \cite{mi2022training}, there is a direct correlation between sensitivity and uncertainty. Sensitivity refers to measuring how a model output changes in response to small perturbations in its input. Mathematically, for a model $f$ with input $\mathbf{x}$ and output $\mathbf{y} = f(\mathbf{x})$, we can define the sensitivity measure as $ S \approx \frac{1}{M} \sum_{i=1}^M \left\| f(P_i(\mathbf{x})) - f(\mathbf{x}) \right\| $
where $P_i(\mathbf{x})$ represents the i-th \textit{perturbed} version of $\mathbf{x}$ according to the scheme P, and $M$ is the number of Monte Carlo samples.
We leverage sensitivity $S$ as a proxy for aleatoric uncertainty $\Ut$ during the diffusion model sampling process for any timestep $t = T \dots 1$. Next, we will define the perturbation scheme.

\paragraph{Perturbation}
A common choice for the perturbation scheme is Gaussian noise, i.e. $P_i(\mathbf{x}) = \mathbf{x} + \epsilon_i$ where $\epsilon_i \sim \mathcal{N}(0, \sigma^2)$.
However, this approach depends on choosing an appropriate noise magnitude $\sigma^2$, which is often non-trivial. To address this limitation, we propose an ad-hoc perturbation scheme specifically designed for diffusion models. Our approach, presented in Sec.~\ref{sec:uncertainty-estimation}, denoises the perturbed image $\Xt$ to obtain the clean image $\Xhatzero$ as in Denoising Diffusion Implicit Model (DDIM) sampler \cite{song2020denoising} and then noise it back to obtain the perturbed image $\Xhatt$.

\subsection{Uncertainty Map Estimation}
\label{sec:method}
\label{sec:uncertainty-estimation}
\label{sec:uncertainty-map-estimation}

We propose to estimate the pixel-wise uncertainty map $\Ut$ during the sampling step $t$ in diffusion models by leveraging the sensitivity of the model output as a proxy for uncertainty estimates.

Let $\Xt$ be the image to be generated at the denoising step $t$, and $\varepsilon_{\theta}(\Xt, t)$ be the score of the image at step $t$. Our algorithm estimates the uncertainty map by first computing an approximation of $\Xzero$ at the current step $t$ as follows:

\begin{equation}
     \Xhatzero = \dfrac{\Xt - \sqrt{1 - \bar{\alpha}_t} \varepsilon_{\theta}(\Xt, t)}{\sqrt{\bar{\alpha}_t}},
\end{equation}
where ${\Xhatzero}$ is an approximation of $\Xzero$ as originally presented in the Denoising Diffusion Implicit Model (DDIM) sampler \cite{song2020denoising}. The approximation ${\Xhatzero}$ is obtained by applying a single denoising step from $\Xt$ to $\Xzero$ using the score $\varepsilon_{\theta}(\Xt, t)$. 

Next, in a Monte-Carlo fashion, we sample $M$ different noisy samples $\left\{ \Xhatt^i:\,\,i=1\dots M \right\}$ from the noising distribution $ q(\Xhatt^i|{\Xhatzero})$ based on Eq.~ \ref{eq:recursive_noising}. This generates $ M $ different versions of $ \Xt $ that are likely to occur as the denoised sample at time step $t$. 
Finally, we compute the uncertainty as the variance of the scores of the generated samples $ \varepsilon_{\theta}(\Xhatt^{i}, t) , i=1\dots M $. The step-wise uncertainty is given by: 
\begin{equation}
    \label{eq:uncertainty_scores}
    \Ut = \text{diag}\left( \left(E_t - \bar{E}_t \right)^T \left(E_t - \bar{E}_t \right)  \right), 
\end{equation}
where \textit{diag} is the diagonal operator, $E_t$ is the tensor obtained by stacking the estimated scores $\left\{ \varepsilontXhatt: i = 1 \dots M \right\}$ $\in \mathbb{R}^{M \times W \times H \times 3}$ and $\bar{E}_t$ the average of $E_t$. Our approach is also illustrated in Fig.~\ref{fig:infer-noise-diff}. By computing the scores $\varepsilon_{\theta}(\Xhatt^{i}, t)$ over $M$ variants of $\Xt$, we identify the most unstable pixels in the denoising step $t$, as the ones with high uncertainty. In this way, our approach can detect artefacts during the generative process. Importantly, we propose an additional interpretation of our uncertainty estimates: the variance of the scores $\varepsilon_{\theta}(\Xhatt^{i}, t)$ can be framed as an approximation of the second order derivative of the noising distribution log-likelihood $\frac{\partial^2}{\partial \Xt^2} \log q(\Xt)$. Next, we explore this relationship in depth in Sec.~\ref{Sec:Fischer} by presenting a detailed analysis of its implications and validity.
\begin{algorithm}
    \caption{Pixel-wise Uncertainty Estimation}
    \label{alg:infer-noise-diff}
    \begin{algorithmic}[1]
    
    \Statex \textbf{Input}: $\Xt$: image at step t, $\bar{\boldsymbol{\alpha}} = \left\{\bar{{\alpha}}_t =  \prod_{s=1}^{t} 1 - \beta_s : t = 1 \dots T \right\}$ where $\beta_s$ is the diffusion noise schedule at timestep $s$, $M$: number of samples for uncertainty estimation
    \Statex \textbf{Output}: The estimated uncertainty
    
    \State Compute true score  $ \varepsilon_{\theta} (\Xt, t)$
    \State ${\Xhatzero} = \dfrac{\Xt - \sqrt{1 - \bar{\alpha}_t} \varepsilon_{\theta}(\Xt, t)}{\sqrt{\bar{\alpha}_t}}$
    \For{$i = 1 \dots M$}
    
        \State $\Xhatt^{i} = \sqrt{\bar{\alpha}_t} {\Xhatzero} + \sqrt{1 - \bar{\alpha}_t} \varepsilon$ \Comment{$\varepsilon \sim N(\mathbf{0}, \mathbf{I})$, Equation \ref{eq:recursive_noising}}
        \State Compute score ${\varepsilon}_{\theta} (\Xhatt^{i}, t)$
    
    \EndFor
    
    \State  $ \Ut = \Var \left( \left\{ \varepsilon_{\theta}(\Xhatt^{i}, t): \,\, i = 1 \dots M \right\} \right) $
    \label{alg:infer-noise-diff:uncertainty}
        
    \Return $\Ut$
    \end{algorithmic}
\end{algorithm}

\subsection{Noising Distribution Curvature}\label{Sec:Fischer}
 
We further explore the relationship between our uncertainty estimates and the second order information of the noising distribution $\frac{\partial^2}{\partial\mathbf{X}_t\partial\mathbf{X}_t^\top} \log q(\Xt)$ for any sampling step of $p_\theta(\Xtm | \Xt) \text{ with } t = 1 \dots T$. We first show the connection between our uncertainty estimation method and the curvature of the marginal noising distribution and then present an intuitive explanation of the uncertainty estimation for the diffusion model.

\paragraph{Connection to the Curvature}

The connection between our uncertainty estimates and the curvature of the noising distribution can be established through the reverse Stochastic Differential Equation (Eq.~\ref{eq:reverse_sde}). It is known that the score approximates the gradient of the noising distribution $\nabla_{\Xt} \log q(\Xt)$ \cite{song2019generative,song2020sliced,song2021scorebased}. Our method, which estimates uncertainty as the variance of the score (Eq.~\ref{eq:uncertainty_scores}), can be related to the second derivative of the noising distribution surface by demonstrating regularity properties similar to those of the Fisher information score \cite{EvansRosenthal2010Book,lehmann_theory_1998}. Detailed proofs and further information on these regularity properties are provided in Appendix A1. Upon establishing the regularity of $\log q(\Xt)$, we arrive at the following relationship, which highlights the connection between our uncertainty estimates and the curvature:

\begin{equation}
    \label{eq:fisher-information-regularity}  
    \begin{split}    
        \mathbf{U}_t & \approx \mathbb{E} \left[ \left(\frac{\partial}{\partial\mathbf{X}_t} \log q(\mathbf{X}_t)\right) \left(\frac{\partial}{\partial\mathbf{X}_t} \log q(\mathbf{X}_t)\right)^\top \right] \\
       & = - \mathbb{E} \left[ \frac{\partial^2}{\partial\mathbf{X}_t\partial\mathbf{X}_t^\top}  \log q (\mathbf{X}_t)  \right].
    \end{split}
\end{equation}

Our uncertainty estimate $\Ut$, as highlighted in Eq. \ref{eq:uncertainty_scores}, approximates the expected value of the second order derivative of the noising distribution $\frac{\partial^2}{\partial\mathbf{X}_t\partial\mathbf{X}_t^\top} \log q(\Xt)$, as we estimate the variance of the scores $\varepsilon_{\theta}(\Xhatt^{i}, t)$ in a Monte-Carlo fashion, using only a subset of the samples. Furthermore, we don't estimate the full variance-covariance matrix, but only the diagonal elements, which are sufficient to provide an estimate of the curvature of the noising distribution.

\paragraph{\textbf{Curvature of q}}
\label{sec:curvature-log-likelihood-fisher-information}
Thanks to the equivalence between the variance of the scores and second derivative, we can interpret the uncertainty estimates as indicators of the curvature of the noising distribution $q(\Xt) = \int p_D(x) q(\Xt | x) dx $.
Therefore, we can leverage the uncertainty estimates to \textit{refine} the generation process, as shown in \cite{filos2020can}. In the next section, we show how to utilise the gradient operation and our uncertainty estimates to guide the sampling process.

\begin{algorithm}
    \caption{Uncertainty Guided Sampling}
    \label{alg:improve-generation}

    \begin{algorithmic}[1]

        \Statex \textbf{Input}: $\XT \sim N(\mathbf{0}, \mathbf{I})$:, $\mathbf{\beta}$: diffusion noise schedule,  $\tauOneT$: Step-wise threhsolds to steer the uncertainty, $M$: number of samples for uncertainty estimation, $\lambda$: strength of the update
        \Statex \textbf{Output}: $ \Xzero $: the generated image
        \For{$ t = T \dots 1 $}
            \State $\varepsilon_t = \varepsilon_{\theta}(\Xt, t)$ \Comment{Compute the score of the image at step t}
            \State $\Ut = \text{uncertainty-estimation}(\Xt, \beta_t, M)$ \Comment{Algorithm \ref{alg:infer-noise-diff}}
            \State $mask = \Ut > \text{percentile}(\Ut, p)$ \Comment{{Compute the mask of the pixels with high uncertainty}}
            \State $\hat{\varepsilon}_t = \varepsilon_t + \lambda (mask \cdot \dfrac{\partial \Ut}{\partial\varepsilon_t})$ \Comment{{Update the score using the gradient of the uncertainty}}
            \State $\Xtm \sim p_\theta(\Xtm | \Xt, \hat{\varepsilon}_t)$ \Comment{Sample from the denoising distribution using the uncertainty guided score}
        \EndFor

    \end{algorithmic}

\end{algorithm}

\subsection{Uncertainty Guided Sampling} 
\label{sec:improving-generation}

Having established the relationship between uncertainty and the second-order derivative of the noising distribution, we propose an algorithm that leverages the uncertainty to guide the sampling process.

To direct the generation, we first establish the high-uncertainty pixels by computing the $p-th$ percentile. Then we compute the uncertainty as highlighted in Alg.~\ref{alg:infer-noise-diff}. Finally, we update the pixels with uncertainty higher than the percentile $p$ using the gradient of the score w.r.t. uncertainty (i.e. gradient ascent) as follows

\begin{equation}
    \hat{\varepsilon}_t = \varepsilon_t + \lambda \left( I[\Ut > p] \cdot \dfrac{\partial\Ut}{\partial\varepsilon_t} \right)
    \label{eq:uncertainty-guided-sampling}
\end{equation}
where $ I[\Ut > p]$ is the indicator function that returns 1 if the pixel has uncertainty higher than percentile p and $\lambda$ the uncertainty update strength.

We guide only high-uncertainty pixels for two reasons. First, we empirically found that high uncertainty pixels are related to foreground elements where most of the artefacts lie. Second, we target pixels with high uncertainty due to our incomplete knowledge of the full noising distribution $q(\Xt)$ so that we are confident to affect most important pixels. Furthermore, this approach not only allows to be applied to unconditional or class-conditional diffusion models \cite{dhariwal2021diffusion,ho2020denoising,song2021scorebased}, but also to text-to-image models like Stable Diffusion \cite{rombach2022high}, as demonstrated in Figure \ref{fig:sd_comparison}.

By explicitly using the uncertainty to guide the sampling, this technique provides a straightforward way to enhance the quality of generations of diffusion models as done in \cite{filos2020can}. But additionally, we provide a theoretical explanation for this in Sec. \ref{Sec:Fischer}. By maximising the uncertainty, we are also maximising the second derivative of the noising distribution (Eq. \ref{eq:fisher-information-regularity}), which is known in literature to improve the convergence rate of optimisation processes \cite{lehmann_theory_1998}.

\section{Experiments}
\label{sec:experiments}

We evaluate our uncertainty estimation and uncertainty guided sampling algorithms in two different settings. First, we filter out low quality image samples and second, we guide the image generation. We also perform an analysis of the generation process and provide visual results on Stable Diffusion \cite{rombach2022high}.

\subsection{Experimental Setup}

\paragraph{Datasets}
We perform evaluation of our method on the ImageNet \cite{deng2009imagenet} dataset, on the variants ImageNet64, ImageNet128, ImageNet256 and ImageNet512 as in BayesDiff \cite{kou2023bayesdiff}. These differ only in the image resolution (respectively, $64 \times 64$, $128 \times 128$, $256 \times 256$, $512 \times 512$). We, additionally, evaluate on the CIFAR-10 dataset using the same protocols.

\paragraph{Models}

We evaluate our approach on the Ablated Diffusion Model (ADM)~\cite{dhariwal2021diffusion}, trained on ImageNet64, and ImageNet128 as well as on the U-ViT model \cite{bao2022all} trained on ImageNet256, ImageNet512. For CIFAR-10, we rely on an open source implementation of the Denoising Diffusion Probabilistic Models (DDPMs) ~\cite{ddpm-cifar10-32} trained on the CIFAR-10 data.  

\paragraph{Evaluation Metrics}

Our evaluation is based on the Fréchet Inception Distance (FID) \cite{heusel2017gans} and well-established uncertainty metrics Area Under the Sparsification Error (AUSE) and Area
Under the Random Gain (AURG). The Fréchet Inception Distance (FID) is a commonly used metric to evaluate the quality and diversity of generated images in generative modelling \cite{NIPS2017_8a1d6947, zhai2020perceptual, chong2020effectively}. It measures the similarity between the distributions of real and generated images by calculating the Fréchet distance between two multivariate Gaussians fitted to feature representations of the Inception-v3 \cite{szegedy2016rethinking} network. Specifically, we take the output of the last pooling layer before the fully connected layers, which has 2048 features similar to BayesDiff \cite{kou2023bayesdiff}. In addition to the FID metric, it is crucial to consider the computational overhead of the uncertainty estimation on top of the diffusion model sampling. To this end, we report the Number of Function Evaluations (NFEs) required by the uncertainty estimation method during the denoising process, as this directly impacts the computational cost and feasibility of the approach in practical scenarios.
Furthermore, we evaluate the uncertainty estimates on the image reconstruction task using AUSE and AURG \cite{ilg2018uncertainty}, both derived from the sparsification plot. This plot is constructed by iteratively removing the pixels with the highest uncertainty from a sample and calculating an error metric at each step. AUSE quantifies the area beneath the sparsification error curve (lower is better). AURG, introduced by \cite{poggi2020uncertainty}, measures the disparity between uncertainty-based sparsification and random sparsification (higher is better).

\paragraph{Evaluation Protocol} 

At first, we create a consistent baseline for each dataset by generating initial points $\XT \sim \Nstandard$ and random labels $y$ using a fixed random seed. This approach ensures a fair comparison across different uncertainty estimation methods by maintaining consistent starting conditions for the denoising process. We evaluate the uncertainty estimation method (Alg. \ref{alg:infer-noise-diff}) and uncertainty guidance method (Alg.~\ref{alg:improve-generation}) with three evaluation protocols. 

To evaluate our uncertainty estimation method, as in BayesDiff, we generate 60,000 images using our diffusion model. From this pool, we create two sets: one comprising 50,000 randomly selected images, and another containing 50,000 images identified as having the highest uncertainty. We then calculate and compare the Fréchet Inception Distance (FID) scores \cite{heusel2017gans} for these two sets. This comparison allows us to quantify the effectiveness of our uncertainty estimation in filtering out low-quality samples and its impact on overall image quality.
Additionally, we evaluate our approach using well-defined uncertainty estimation metrics \cite{ilg2018uncertainty, poggi2020uncertainty} using the evaluation protocol of AnoDDPM \cite{Wyatt_2022_CVPR}. We sample ground-truth test images and we inject noise as defined in Sec.~\ref{par:noising} until half of the noising process (i.e. $T/2$). Then, we denoise the images using the diffusion model and compute the reconstruction error using the Root Mean Squared Error (RMSE). Finally, we compute the sparsification error curve, and consequently AUSE and AURG metrics, using the uncertainty computed during the sampling process. Finally, for the uncertainty guidance evaluation, we generate $ 10\,000 $ images with and without the uncertainty guidance from the same diffusion model to compare the FID score.

\paragraph{Comparisons}

We compare our uncertainty estimation method with the model-agnostic uncertainty estimation method MC-Dropout \cite{gal2016dropout}, which is applied to the ADM model, trained on ImageNet64 \cite{deng2009imagenet} and CIFAR-10 \cite{Krizhevsky09learningmultipleCifar10}. In addition, we compare our method with BayesDiff \cite{kou2023bayesdiff}, which also performs uncertainty estimation.

\paragraph{Implementation Details}

We generate all samples using the samplers DDIM~\cite{song2020denoising} and second-order DPM~\cite{lu2022dpm} with 50 generation steps. We set the number of estimated scores $M$ to $5$ for the uncertainty estimation. We compute the uncertainty of the generated image by summing the pixel-wise uncertainty from denoising timestep 45 until 48. For the uncertainty-guided generation, we compute the threshold value as the 95-th percentile of the uncertainty computed over the $ 10\,000 $ samples generated from the diffusion model with strength $\lambda = 1.0$. 

\subsection{Result Discussion}

\paragraph{Uncertainty Estimation}

We present our uncertainty estimation results in Table \ref{tab:filtering-results}. While all approaches improve with respect to the random baselines, we deliver the best FID score in all the cases except for ImageNet256. Also, our approach demonstrates enhanced computational efficiency with a total of 20 Number of Function Evaluations (NFEs), compared to approximately 130 NFEs required by BayesDiff for 50-step generations \cite{kou2023bayesdiff}. This is due to the uncertainty schedule described in Fig.~\ref{fig:uncertainty-schedule-timesteps} which exhibits high variability of the uncertainty during the last few generation steps. Finally, our method requires fewer NFEs (50) than MC-Dropout.
In the image reconstruction task, we achieve a lower AUSE and higher AURG score compared to MC-Dropout, as shown in Table \ref{tab:ause_aurg_comparison}. As shown in Figure 1 and 2 in the Appendix, 
we show that the uncertainty computed by MC-Dropout does not capture the uncertainty of the data distribution as effectively as our method.



\begin{table*}[t]
    \caption{Comparison of the FID score between $ 60\,000 $ generated images with and without the uncertainty guidance. The missing results of BayesDiff are not available, while the missing results from MC-Dropout are not computable as the model is not trained with Dropout enabled. Random baseline comes from our experiments.}
    \vspace{-.1cm}
    \label{tab:filtering-results}
    \setlength{\tabcolsep}{0.55em} %
    \small
    \centering
    \begin{tabular}{ccrrrr}
    \toprule
    \multirow{2}{*}{Model}&\multirow{2}{*}{Dataset}& \multicolumn{4}{c}{\text{FID} $\downarrow$} \\
    \cmidrule{3-6} && Random & Ours & BayesDiff & MC-Dropout \\
    \midrule
    ADM&ImageNet 64& 3.289 & \textbf{3.254} & - & 3.268 \\
    ADM&ImageNet 128& 8.21 & \textbf{7.88} & 8.45 & - \\
    ADM w/2-DPM & ImageNet 128 & 8.50 & \textbf{8.48} & 9.67 & - \\
    U-ViT&ImageNet 256& 7.88 & 7.80 & \textbf{6.81} & - \\
    U-ViT&ImageNet 512& 16.47 & \textbf{16.37} & 16.87 & - \\
    DDPM & CIFAR-10 & 13.494 & \textbf{13.416} & - & 13.435 \\
    \bottomrule
    \end{tabular}
    \vspace{-.2cm}
\end{table*}
\setblue
\begin{table}[t]
    \caption{Comparison of AUSE$\downarrow$/AURG$\uparrow$ scores for Our Method and MC-Dropout on ImageNet64 and CIFAR-10 datasets.}
    \vspace{-.1cm}
    \label{tab:ause_aurg_comparison}
    \setlength{\tabcolsep}{0.55em}
    \small
    \centering
    \begin{tabular}{crr}
    \toprule
    \multirow{2}{*}{Dataset} & \multicolumn{2}{c}{AUSE$\downarrow$/AURG$\uparrow$} \\
    \cmidrule(lr){2-3}
    & Our Method & MC-Dropout \\
    \midrule
    ImageNet64 & \textbf{74.48/5.05} & $84.94$/$-4.85$ \\
    CIFAR-10 & \textbf{0.01/18.48} &  $1.27$/$16.19$ \\
    \bottomrule
    \end{tabular}
    \vspace{-.2cm}
\end{table}
\setblack

\begin{table}[t]
    \caption{Comparison of the FID score between $ 10\,000 $ generated images with and without the uncertainty guidance.}
    \vspace{-.1cm}
    \label{tab:uncertainty-guidance-results}
    \setlength{\tabcolsep}{0.55em} %
    \small
    \centering
    \begin{tabular}{ccrr}
    \toprule
    \multirow{2}{*}{Model}&\multirow{2}{*}{Dataset}& \multicolumn{2}{c}{\text{FID} $\downarrow$} \\
    \cmidrule{3-4} && normal & uncertainty guided \\
    \midrule
    ADM&ImageNet 64& 24.16 & \textbf{23.21} \\
    ADM&ImageNet 128& 45.10 & \textbf{44.02} \\
    DDPM&CIFAR-10& 27.39 & \textbf{26.45} \\
    U-ViT&ImageNet 256&  51.45 & \textbf{50.34} \\
    U-ViT&ImageNet 512&  60.72 & \textbf{59.81} \\
    \bottomrule
    \end{tabular}
    \vspace{-.2cm}
\end{table}

\paragraph{Uncertainty Guided Sampling}

In Table \ref{tab:uncertainty-guidance-results}, we compare the FID score of images generated with and without the uncertainty guidance, using the same initial points $ \XT $. We observe a clear improvement of $\approx 1$ when the uncertainty guidance is employed on the same set of images. This can be considered as empirical evidence that the uncertainty computed by our method not only can detect low quality samples but can also utilise the uncertainty to steer the denoising process toward higher quality images. 

\paragraph{Qualitative analysis}

Fig.~\ref{fig:sd_comparison} illustrates visual results of our approach when applied to Stable Diffusion \cite{rombach2022high}. We can see that uncertainty guided images have less unrealistic artefacts or no artefacts. In addition, they usually contain more contextual details compared to the generated images without uncertainty guidance. For instance, the last column of 


\subsection{Further Analysis}

\begin{figure}
    \centering
    \includegraphics[width=0.48\linewidth]{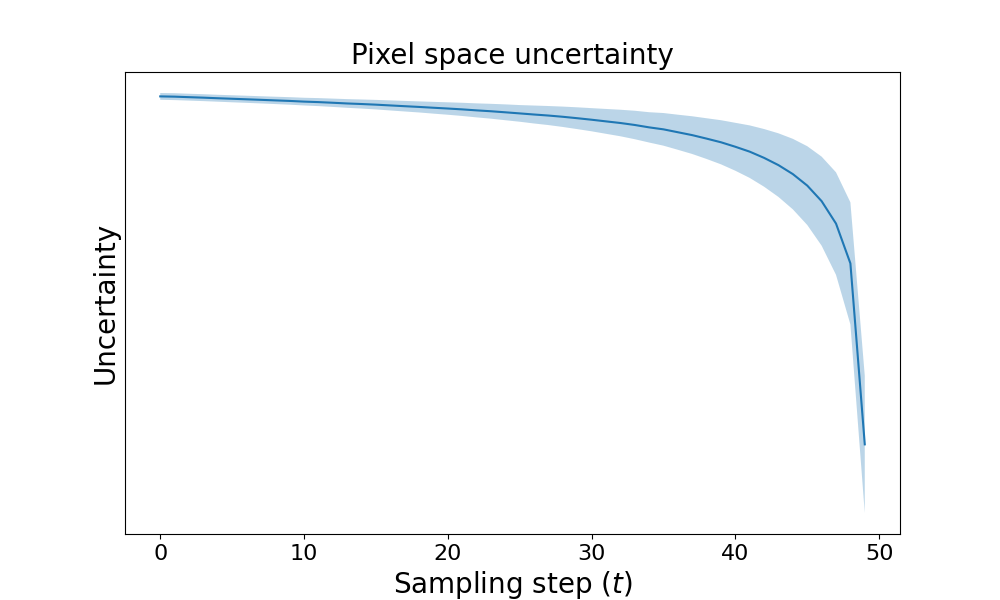}
    \includegraphics[width=0.48\linewidth]{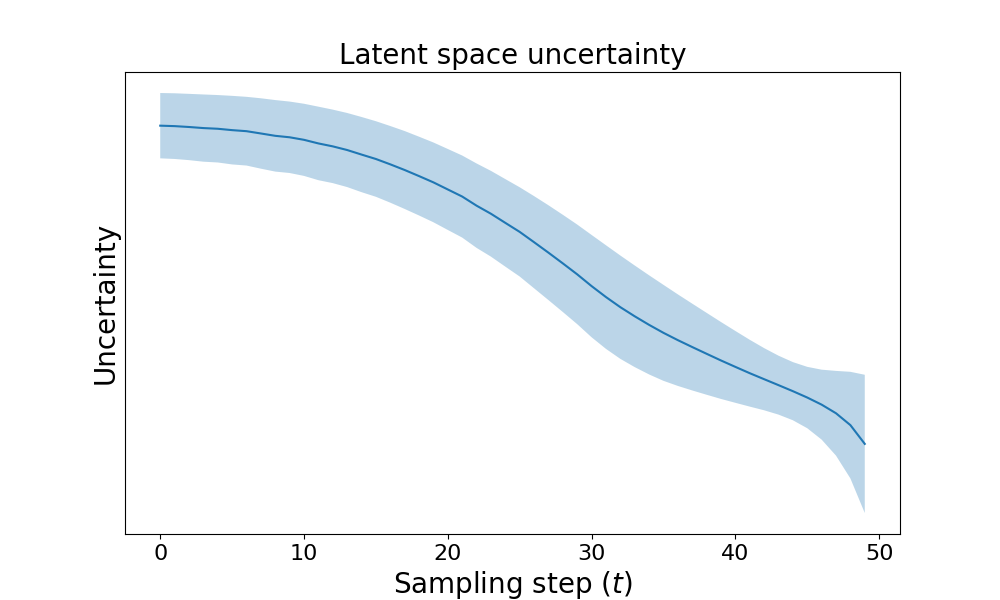}
    \caption{We present posterior uncertainty in pixel (left) and latent (right) spaces. The blue line shows average uncertainty over 60,000 samples, with standard deviation in the surrounding blue area. This pattern was consistent across all evaluated models. 
    }
    \label{fig:uncertainty-maps}
    \label{fig:uncertainty-schedule-timesteps}
\end{figure}

Next, we analyse the variance of the step-wise uncertainty, i.e.~uncertainty for each diffusion sampling step, for over $ 60\,000 $ generated samples using ADM on ImageNet64 to gain insights about the relation between uncertainty and the denoising process. Fig.~\ref{fig:uncertainty-schedule-timesteps}, in pixel space, highlights a high variability in uncertainty during the final stages of the diffusion process, particularly between 75\% and 90\% of the denoising process, while remaining relatively stable throughout the rest of the process. This trend can be attributed to the model determining foreground elements in the later stages of the sampling process. 

\section{Conclusion}
We presented an approach for pixel-wise uncertainty estimation during the sampling phase of diffusion models. For each sampling step, we estimated the uncertainty as the variance of the denoising values using multiple generated samples. We then demonstrated the relationship between the uncertainty estimates and the second derivative of the log-likelihood of the noising distribution. Based on this connection, we presented an algorithm to guide the sampling phase of diffusion models. By guiding the sampling process with our uncertainty estimates, we achieve better image quality. In our evaluations, we show that our uncertainty estimation approach filters out low quality samples generated by the diffusion models, such as ADM and U-VIT trained on ImageNet and CIFAR-10. We also show that uncertainty-guided sampling improves the quality of the generated samples using the FID score as a metric. Furthermore, our approach outperformed the related work in almost all evaluations.

\section{Acknowledgements}
\label{sec:acknowledgements}
Part of the research leading to these results is funded by the German Research Foundation (DFG) within the project \textit{Transferring Deep Neural Networks from Simulation to Real-World} (project number 458972748). The authors would like to thank the foundation for the successful cooperation. Additionally the authors gratefully acknowledge the scientific support and HPC resources provided by the Erlangen National High Performance Computing Center (NHR@FAU) of the Friedrich-Alexander-Universität Erlangen-Nürnberg (FAU). The hardware is funded by the German Research Foundation (DFG).

M.D.V. thanks Giovanni Barbarani and Rohan Asthana for the helpful discussions and support.

\setblack

%
%
\bibliographystyle{splncs04}
\bibliography{main}

\begin{thebibliography}{10}
\providecommand{\url}[1]{\texttt{#1}}
\providecommand{\urlprefix}{URL }
\providecommand{\doi}[1]{https://doi.org/#1}

\bibitem{An2015VariationalAB}
An, J., Cho, S.: Variational autoencoder based anomaly detection using reconstruction probability (2015), \url{https://api.semanticscholar.org/CorpusID:36663713}

\bibitem{anderson_reverse-time_1982}
Anderson, B.D.O.: Reverse-time diffusion equation models. Stochastic Processes and their Applications  \textbf{12}(3),  313--326 (1982). \doi{https://doi.org/10.1016/0304-4149(82)90051-5}, \url{https://www.sciencedirect.com/science/article/pii/0304414982900515}

\bibitem{asthana2024multiconditioned}
Asthana, R., Conrad, J., Dawoud, Y., Ortmanns, M., Belagiannis, V.: Multi-conditioned graph diffusion for neural architecture search. Transactions on Machine Learning Research  (2024), \url{https://openreview.net/forum?id=5VotySkajV}

\bibitem{bao2022all}
Bao, F., Li, C., Cao, Y., Zhu, J.: All are worth words: a vit backbone for score-based diffusion models. In: NeurIPS 2022 Workshop on Score-Based Methods (2022), \url{https://openreview.net/forum?id=WfkBiPO5dsG}

\bibitem{pmlr-v37-blundell15}
Blundell, C., Cornebise, J., Kavukcuoglu, K., Wierstra, D.: Weight uncertainty in neural network. In: Bach, F., Blei, D. (eds.) Proceedings of the 32nd International Conference on Machine Learning. Proceedings of Machine Learning Research, vol.~37, pp. 1613--1622. PMLR, Lille, France (07--09 Jul 2015), \url{https://proceedings.mlr.press/v37/blundell15.html}

\bibitem{chen2018deep}
Chen, X., Pawlowski, N., Rajchl, M., Glocker, B., Konukoglu, E.: Deep generative models in the real-world: An open challenge from medical imaging. arXiv preprint arXiv:1806.05452  (2018)

\bibitem{chong2020effectively}
Chong, M.J., Forsyth, D.: Effectively unbiased fid and inception score and where to find them. In: Proceedings of the IEEE/CVF conference on computer vision and pattern recognition. pp. 6070--6079 (2020)

\bibitem{laplace2021}
Daxberger, E., Kristiadi, A., Immer, A., Eschenhagen, R., Bauer, M., Hennig, P.: Laplace redux--effortless {B}ayesian deep learning. In: {N}eur{IPS} (2021)

\bibitem{deng2009imagenet}
Deng, J., Dong, W., Socher, R., Li, L.J., Li, K., Fei-Fei, L.: Imagenet: A large-scale hierarchical image database. In: 2009 IEEE conference on computer vision and pattern recognition. pp. 248--255. Ieee (2009)

\bibitem{dhariwal2021diffusion}
Dhariwal, P., Nichol, A.: Diffusion models beat gans on image synthesis. Advances in neural information processing systems  \textbf{34},  8780--8794 (2021)

\bibitem{esser2024scaling}
Esser, P., Kulal, S., Blattmann, A., Entezari, R., M{\"u}ller, J., Saini, H., Levi, Y., Lorenz, D., Sauer, A., Boesel, F., et~al.: Scaling rectified flow transformers for high-resolution image synthesis. In: Forty-first International Conference on Machine Learning

\bibitem{EvansRosenthal2010Book}
Evans, M.J., Rosenthal, J.S.: Probability and Statistics: The Science of Uncertainty. University of Toronto, 2 edn. (2010)

\bibitem{filos2020can}
Filos, A., Tigkas, P., McAllister, R., Rhinehart, N., Levine, S., Gal, Y.: Can autonomous vehicles identify, recover from, and adapt to distribution shifts? In: International Conference on Machine Learning. pp. 3145--3153. PMLR (2020)

\bibitem{gal2016dropout}
Gal, Y., Ghahramani, Z.: Dropout as a bayesian approximation: Representing model uncertainty in deep learning. In: international conference on machine learning. pp. 1050--1059. PMLR (2016)

\bibitem{goodfellow2020generative}
Goodfellow, I., Pouget-Abadie, J., Mirza, M., Xu, B., Warde-Farley, D., Ozair, S., Courville, A., Bengio, Y.: Generative adversarial networks. Communications of the ACM  \textbf{63}(11),  139--144 (2020)

\bibitem{ddpm-cifar10-32}
Google: Denoising diffusion probabilistic model (ddpm) trained on cifar-10 at 32x32 resolution. \url{https://huggingface.co/google/ddpm-cifar10-32} (2022)

\bibitem{pmlr-v89-grover19a}
Grover, A., Ermon, S.: Uncertainty autoencoders: Learning compressed representations via variational information maximization. In: Chaudhuri, K., Sugiyama, M. (eds.) Proceedings of the Twenty-Second International Conference on Artificial Intelligence and Statistics. Proceedings of Machine Learning Research, vol.~89, pp. 2514--2524. PMLR (16--18 Apr 2019), \url{https://proceedings.mlr.press/v89/grover19a.html}

\bibitem{hemsley2020deep}
Hemsley, M., Chugh, B., Ruschin, M., Lee, Y., Tseng, C.L., Stanisz, G., Lau, A.: Deep generative model for synthetic-ct generation with uncertainty predictions. In: Medical Image Computing and Computer Assisted Intervention--MICCAI 2020: 23rd International Conference, Lima, Peru, October 4--8, 2020, Proceedings, Part I 23. pp. 834--844. Springer (2020)

\bibitem{heusel2017gans}
Heusel, M., Ramsauer, H., Unterthiner, T., Nessler, B., Hochreiter, S.: Gans trained by a two time-scale update rule converge to a local nash equilibrium. Advances in neural information processing systems  \textbf{30} (2017)

\bibitem{NIPS2017_8a1d6947}
Heusel, M., Ramsauer, H., Unterthiner, T., Nessler, B., Hochreiter, S.: Gans trained by a two time-scale update rule converge to a local nash equilibrium. In: Guyon, I., Luxburg, U.V., Bengio, S., Wallach, H., Fergus, R., Vishwanathan, S., Garnett, R. (eds.) Advances in Neural Information Processing Systems. vol.~30. Curran Associates, Inc. (2017), \url{https://proceedings.neurips.cc/paper_files/paper/2017/file/8a1d694707eb0fefe65871369074926d-Paper.pdf}

\bibitem{ho2020denoising}
Ho, J., Jain, A., Abbeel, P.: Denoising diffusion probabilistic models. Advances in neural information processing systems  \textbf{33},  6840--6851 (2020)

\bibitem{hornauer2022gradient}
Hornauer, J., Belagiannis, V.: Gradient-based uncertainty for monocular depth estimation. In: European Conference on Computer Vision. pp. 613--630. Springer (2022)

\bibitem{hornauer2023heatmap}
Hornauer, J., Belagiannis, V.: Heatmap-based out-of-distribution detection. In: Proceedings of the IEEE/CVF Winter Conference on Applications of Computer Vision. pp. 2603--2612 (2023)

\bibitem{hornauer2023out}
Hornauer, J., Holzbock, A., Belagiannis, V.: Out-of-distribution detection for monocular depth estimation. In: Proceedings of the IEEE/CVF International Conference on Computer Vision. pp. 1911--1921 (2023)

\bibitem{huang2017snapshot}
Huang, G., Li, Y., Pleiss, G., Liu, Z., Hopcroft, J.E., Weinberger, K.Q.: Snapshot ensembles: Train 1, get m for free. arXiv preprint arXiv:1704.00109  (2017)

\bibitem{ilg2018uncertainty}
Ilg, E., Cicek, O., Galesso, S., Klein, A., Makansi, O., Hutter, F., Brox, T.: Uncertainty estimates and multi-hypotheses networks for optical flow. In: Proceedings of the European Conference on Computer Vision (ECCV). pp. 652--667 (2018)

\bibitem{ioffe2015batch}
Ioffe, S., Szegedy, C.: Batch normalization: Accelerating deep network training by reducing internal covariate shift. In: International conference on machine learning. pp. 448--456. pmlr (2015)

\bibitem{kingma2021variational}
Kingma, D., Salimans, T., Poole, B., Ho, J.: Variational diffusion models. Advances in neural information processing systems  \textbf{34},  21696--21707 (2021)

\bibitem{kingma2013auto}
Kingma, D.P., Welling, M.: Auto-encoding variational bayes  (2013)

\bibitem{kou2023bayesdiff}
Kou, S., Gan, L., Wang, D., Li, C., Deng, Z.: Bayesdiff: Estimating pixel-wise uncertainty in diffusion via bayesian inference. arXiv preprint arXiv:2310.11142  (2023)

\bibitem{Krizhevsky09learningmultipleCifar10}
Krizhevsky, A.: Learning multiple layers of features from tiny images. Tech. rep. (2009)

\bibitem{lakshminarayanan2017simple}
Lakshminarayanan, B., Pritzel, A., Blundell, C.: Simple and scalable predictive uncertainty estimation using deep ensembles. Advances in neural information processing systems  \textbf{30} (2017)

\bibitem{lehmann_theory_1998}
Lehmann, E.L., Casella, G.: Theory of point estimation. Springer texts in statistics, Springer, New York, NY, 2. ed edn. (1998)

\bibitem{lu2022dpm}
Lu, C., Zhou, Y., Bao, F., Chen, J., Li, C., Zhu, J.: Dpm-solver++: Fast solver for guided sampling of diffusion probabilistic models  (2022)

\bibitem{mi2022training}
Mi, L., Wang, H., Tian, Y., He, H., Shavit, N.N.: Training-free uncertainty estimation for dense regression: Sensitivity as a surrogate. In: Proceedings of the AAAI Conference on Artificial Intelligence. vol.~36, pp. 10042--10050 (2022)

\bibitem{morales2020activation}
Morales-Alvarez, P., Hern{\'a}ndez-Lobato, D., Molina, R., Hern{\'a}ndez-Lobato, J.M.: Activation-level uncertainty in deep neural networks. In: International Conference on Learning Representations (2020)

\bibitem{neumeier2024reliable}
Neumeier, M., Dorn, S., Botsch, M., Utschick, W.: Reliable trajectory prediction and uncertainty quantification with conditioned diffusion models. In: Proceedings of the IEEE/CVF Conference on Computer Vision and Pattern Recognition. pp. 3461--3470 (2024)

\bibitem{nielsen2024diffenc}
Nielsen, B.M.G., Christensen, A., Dittadi, A., Winther, O.: Diffenc: Variational diffusion with a learned encoder. In: The Twelfth International Conference on Learning Representations (2024), \url{https://openreview.net/forum?id=8nxy1bQWTG}

\bibitem{notin2021improving}
Notin, P., Hern{\'a}ndez-Lobato, J.M., Gal, Y.: Improving black-box optimization in vae latent space using decoder uncertainty. Advances in Neural Information Processing Systems  \textbf{34},  802--814 (2021)

\bibitem{poggi2020uncertainty}
Poggi, M., Aleotti, F., Tosi, F., Mattoccia, S.: On the uncertainty of self-supervised monocular depth estimation. In: Proceedings of the IEEE/CVF Conference on Computer Vision and Pattern Recognition. pp. 3227--3237 (2020)

\bibitem{rombach2022high}
Rombach, R., Blattmann, A., Lorenz, D., Esser, P., Ommer, B.: High-resolution image synthesis with latent diffusion models. In: Proceedings of the IEEE/CVF conference on computer vision and pattern recognition. pp. 10684--10695 (2022)

\bibitem{saatci2017bayesian}
Saatci, Y., Wilson, A.G.: Bayesian gan. Advances in neural information processing systems  \textbf{30} (2017)

\bibitem{sagar2022uncertainty}
Sagar, A.: Uncertainty quantification using variational inference for biomedical image segmentation. In: Proceedings of the IEEE/CVF Winter Conference on Applications of Computer Vision. pp. 44--51 (2022)

\bibitem{song2020denoising}
Song, J., Meng, C., Ermon, S.: Denoising diffusion implicit models. In: International Conference on Learning Representations (2020)

\bibitem{song2019generative}
Song, Y., Ermon, S.: Generative modeling by estimating gradients of the data distribution. Advances in neural information processing systems  \textbf{32} (2019)

\bibitem{song2020sliced}
Song, Y., Garg, S., Shi, J., Ermon, S.: Sliced score matching: A scalable approach to density and score estimation. In: Uncertainty in Artificial Intelligence. pp. 574--584. PMLR (2020)

\bibitem{song2021scorebased}
Song, Y., Sohl-Dickstein, J., Kingma, D.P., Kumar, A., Ermon, S., Poole, B.: Score-based generative modeling through stochastic differential equations. In: International Conference on Learning Representations (2021), \url{https://openreview.net/forum?id=PxTIG12RRHS}

\bibitem{szegedy2016rethinking}
Szegedy, C., Vanhoucke, V., Ioffe, S., Shlens, J., Wojna, Z.: Rethinking the inception architecture for computer vision. In: Proceedings of the IEEE conference on computer vision and pattern recognition. pp. 2818--2826 (2016)

\bibitem{pmlr-v80-teye18a}
Teye, M., Azizpour, H., Smith, K.: {B}ayesian uncertainty estimation for batch normalized deep networks. In: Dy, J., Krause, A. (eds.) Proceedings of the 35th International Conference on Machine Learning. Proceedings of Machine Learning Research, vol.~80, pp. 4907--4916. PMLR (10--15 Jul 2018), \url{https://proceedings.mlr.press/v80/teye18a.html}

\bibitem{wiederer2023joint}
Wiederer, J., Schmidt, J., Kressel, U., Dietmayer, K., Belagiannis, V.: Joint out-of-distribution detection and uncertainty estimation for trajectory prediction. In: 2023 IEEE/RSJ International Conference on Intelligent Robots and Systems (IROS). pp. 5487--5494. IEEE (2023)

\bibitem{wilson2020bayesian}
Wilson, A.G., Izmailov, P.: Bayesian deep learning and a probabilistic perspective of generalization. Advances in neural information processing systems  \textbf{33},  4697--4708 (2020)

\bibitem{Wyatt_2022_CVPR}
Wyatt, J., Leach, A., Schmon, S.M., Willcocks, C.G.: Anoddpm: Anomaly detection with denoising diffusion probabilistic models using simplex noise. In: Proceedings of the IEEE/CVF Conference on Computer Vision and Pattern Recognition (CVPR) Workshops. pp. 650--656 (June 2022)

\bibitem{zhai2020perceptual}
Zhai, G., Min, X.: Perceptual image quality assessment: a survey. Science China Information Sciences  \textbf{63},  1--52 (2020)

\end{thebibliography}

\appendix
\newpage
\section{Proof of the main statement}

  In this section we provide the proof that the expected value of the variance of the scores is equivalent to the expected value of the second derivative of the noising distribution $q_t(\Xt) = \int p(\mathbf{X}) q_t(\Xt|\mathbf{X}) dX$
  \begin{eqnarray}
      \label{eq:proof}
        &  \mathbb{E} & \left[ \left(\frac{\partial}{\partial \Xt } \log q( \Xt )\right) \left(\frac{\partial}{\partial \Xt } \log q( \Xt )\right)^\top \right] = \\
        & = & - \mathbb{E} \left[ \frac{\partial^2}{\partial \Xt \partial \Xt ^\top}  \log q ( \Xt )  \right].
 \end{eqnarray}

  In the main text we use this result to gain insight about our uncertainty estimates, which approximate the expected value of the variance of the scores with a Monte Carlo estimate i.e.
  \begin{eqnarray}
      \label{eq:mc_estimate}
      \Ut & = & \text{diag}\left( \left(E_t - \bar{E}_t \right)^T \left(E_t - \bar{E}_t \right)  \right) \\
      & \approx & \mathbb{E} \left[ \left(\frac{\partial}{\partial \Xt } \log q( \Xt )\right) \left(\frac{\partial}{\partial \Xt } \log q( \Xt )\right)^\top \right]
    \end{eqnarray}

  where "diag" is the diagonal operator, $E_t$ is the matrix obtained by stacking the estimated scores $\left\{ \varepsilontXhatt: i = 1 \dots M \right\}$ and $\bar{E}_t$ the average of $E_t$.
  
  Now we provide the proof of Eq. \ref{eq:proof}. For the sake of simplicity, we demonstrate our statement for a scalar x 
\paragraph{Theorem}

Suppose that response $x$ is real-valued, and the noising distribution $q(x)$ satisfies the following regularity conditions:

\setblue
\begin{equation}
q(x) \in C^2
\label{eq:condition1}
\end{equation}

i.e. q(x) is twice continuously differentiable and

\setblack

\begin{equation}
\int_{-\infty}^{\infty} \left| \frac{\partial^2 \log q(x)}{\partial x^2} \right| q(x) dx < \infty
\label{eq:condition4}
\end{equation}

Then we have the main result:
\begin{equation}
  \begin{split}
      \label{eq:main_statement}
    &  \mathbb{E}  \left[ \left(\frac{\partial}{\partial x } \log q( x )\right)^2 \right] = \\
    & =  - \mathbb{E} \left[ \frac{\partial^2}{\partial x^2 }  \log q ( x )  \right].
  \end{split}
\end{equation}
\paragraph{Proof}

To prove that LHS = RHS, we can start with the right-hand side and show that it equals the left-hand side.

\begin{enumerate}

    \item First, we expand the RHS:
    \begin{equation}
      - \mathbb{E} \left[ \frac{\partial^2}{\partial x^2 }  \log q ( x )  \right] = - \int q(x) \frac{\partial^2}{\partial x^2 }  \log q ( x ) dx
    \end{equation}

    \item Using the chain rule:

    \begin{equation}
        \frac{\partial^2}{\partial x^2 }  \log q ( x ) = \frac{\partial}{\partial x} \left(\frac{1}{q(x)} \frac{\partial q(x)}{\partial x}\right)
    \end{equation}

    Then by applying the product rule for differentiation, which states that $(u \cdot v)' = u \cdot v' + v \cdot u'$ we have that

    \begin{equation}
        = -\frac{1}{q(x)^2} \left(\frac{\partial q(x)}{\partial x}\right)^2 + \frac{1}{q(x)} \frac{\partial^2 q(x)}{\partial x^2}
    \end{equation}

    \item Substituting this back into the integral:

    $$- \int q(x) \left(-\frac{1}{q(x)^2} \left(\frac{\partial q(x)}{\partial x}\right)^2 + \frac{1}{q(x)} \frac{\partial^2 q(x)}{\partial x^2}\right) dx$$
    $$= \int \frac{1}{q(x)} \left(\frac{\partial q(x)}{\partial x}\right)^2 dx - \int \frac{\partial^2 q(x)}{\partial x^2} dx$$

    \setblue
    \item The second term becomes zero due to the property in Eq. \ref{eq:condition1} as:
    
    \begin{equation}
      \int \frac{\partial^2 q(x)}{\partial x^2} dx = \frac{\partial q(x)}{\partial x} |_{-\infty}^{\infty}
    \end{equation}

    Finally, considering that q(x) is a probability distribution, its derivative $\frac{\partial q(x)}{\partial x}$ is 0 when diverging to $\pm \infty$, hence

    \begin{equation}
      \frac{\partial q(x)}{\partial x} |_{-\infty}^{\infty} = 0
    \end{equation}

    Now, going back to the first term



    \setblack
    \begin{equation}
      \int \frac{1}{q(x)} \left(\frac{\partial q(x)}{\partial x}\right)^2 dx 
    \end{equation}

    \item We can multiply and divide the integrand by $q(x)$ without changing the value of the integral:

    \begin{equation}
      \int \frac{q(x)}{q(x)} \left(\frac{\partial q(x)}{\partial x}\right)^2 \frac{1}{q(x)} dx
    \end{equation}

    \item This can be rewritten as:

    \begin{equation}
      \int q(x) \left(\frac{1}{q(x)} \frac{\partial q(x)}{\partial x}\right)^2 dx   
    \end{equation}

    \item Now, we can use the following identity:
    \begin{equation}
      \frac{1}{q(x)} \frac{\partial q(x)}{\partial x} = \frac{\partial \log q(x)}{\partial x}
    \end{equation}

    \item Substituting this identity into the previous expression, we get:

    \begin{equation}
      \int q(x) \left(\frac{\partial \log q(x)}{\partial x}\right)^2 dx
    \end{equation}
        
    \item This is exactly the definition of the left-hand side of the original equation:

    \begin{equation}
      \mathbb{E}  \left[ \left(\frac{\partial}{\partial x } \log q( x )\right)^2 \right]
    \end{equation}

    Therefore, we have shown that the right-hand side equals the left-hand side, proving the identity.
\end{enumerate}

\section{Additional figures}

\begin{center}
  \begin{figure}[H]
    \centering
    \includegraphics[width=0.9\linewidth]{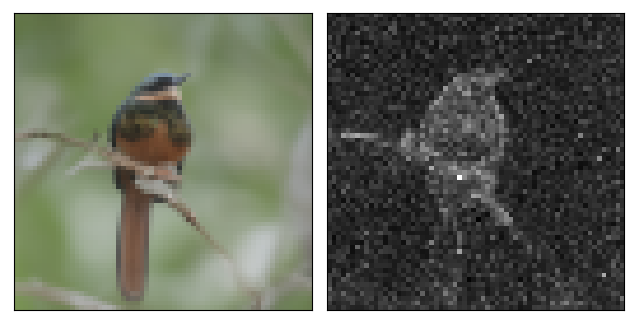}
    \caption{Left: generated image from DDPM trained on Imagenet64 with 50 steps and DDIM sampler. Right: uncertainty map of the generated image. The uncertainty map is obtained by summing the step-wise uncertainty of the sampling process. We observe that most of the uncertainty is concentrated in the foreground elements of the image.}
    \label{fig:uncertainty}
  \end{figure}
\end{center}

\begin{figure}[H]
    \centering
    \includegraphics[width=0.9\linewidth]{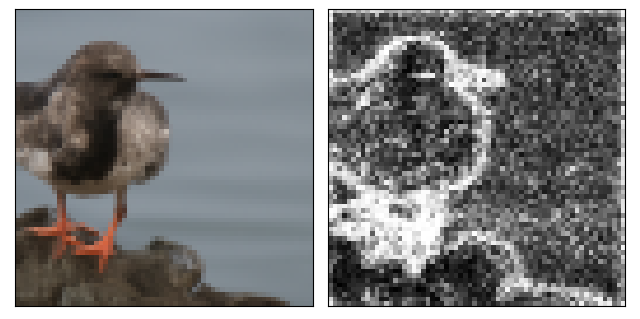}
    \caption{Left: generated image from DDPM trained on Imagenet64 with 50 steps and DDIM sampler. Right: uncertainty map from MC-Dropout of the generated image. The uncertainty map is obtained by summing the step-wise uncertainty of the sampling process. We observe that most of the uncertainty is concentrated in the edges of the foreground elements of the image.}
\end{figure}

\begin{figure*}
  \centering
  \includegraphics[width=0.85\linewidth]{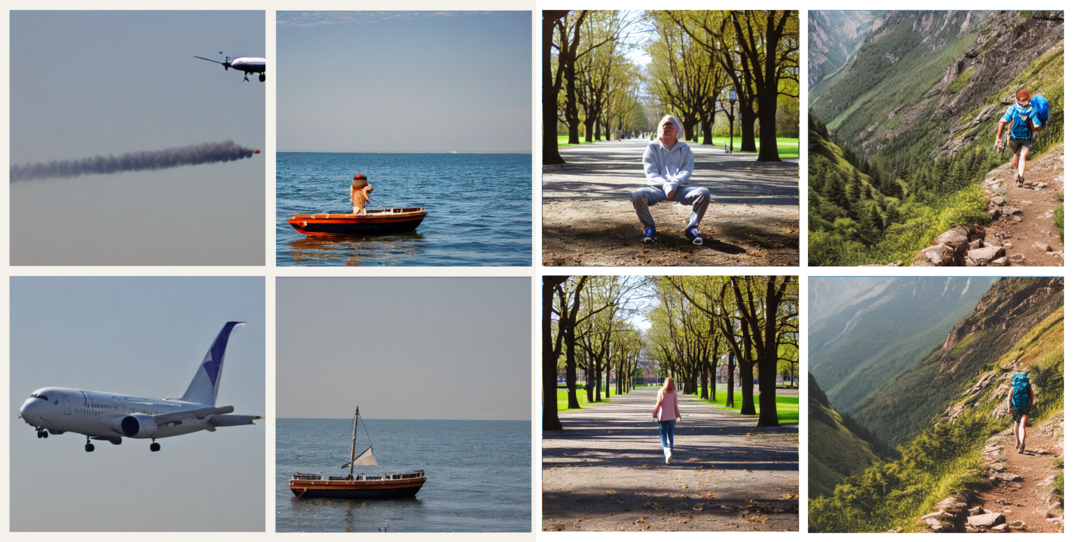}
\caption{Additional visual results of uncertainty guidance applied to Stable Diffusion. For each pair of images, the \textit{top row} shows the generated image \textit{without} uncertainty guidance while the \textbf{bottom row} shows the same image generated \textbf{with} uncertainty guidance.}
\end{figure*}

\begin{figure*}
  \centering
  \includegraphics[width=0.85\linewidth]{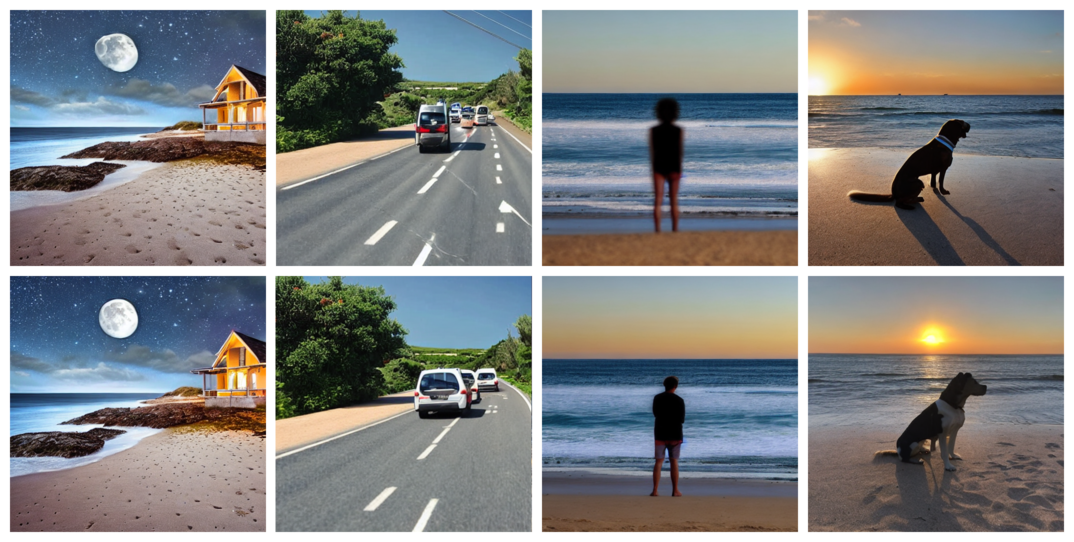}
\caption{Additional visual results of uncertainty guidance applied to Stable Diffusion. For each pair of images, the \textit{top row} shows the generated image \textit{without} uncertainty guidance while the \textbf{bottom row} shows the same image generated \textbf{with} uncertainty guidance.}
\end{figure*}

\begin{figure*}
  \includegraphics[width=0.99\linewidth]{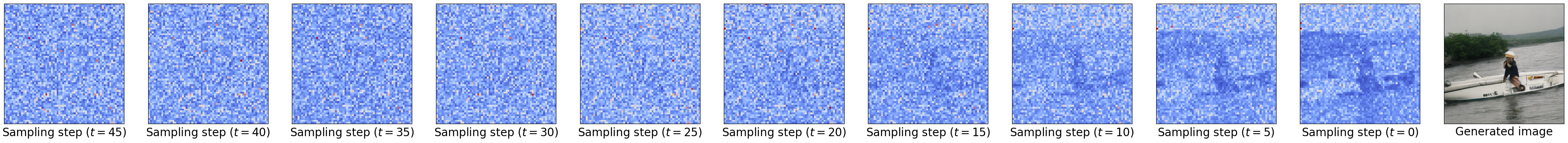}
  \includegraphics[width=0.99\linewidth]{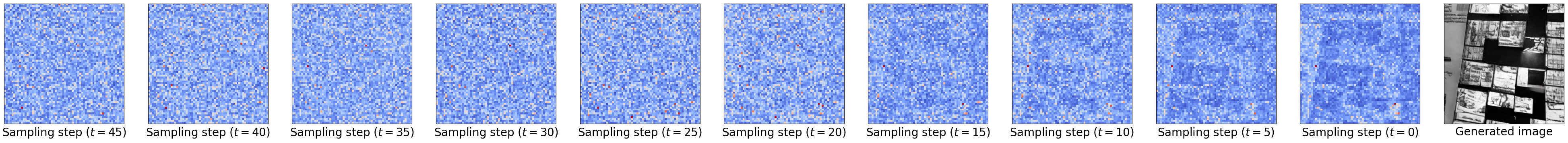}
  \includegraphics[width=0.99\linewidth]{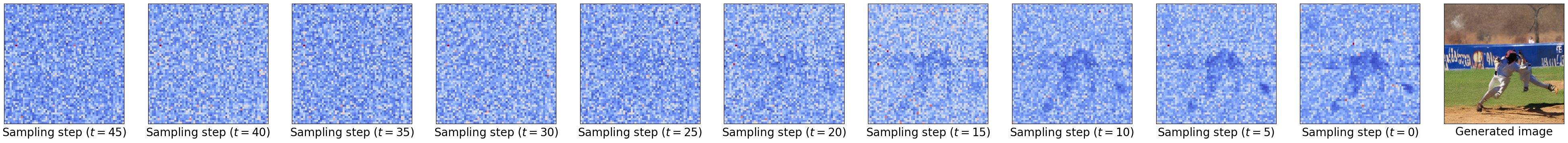}
  \includegraphics[width=0.99\linewidth]{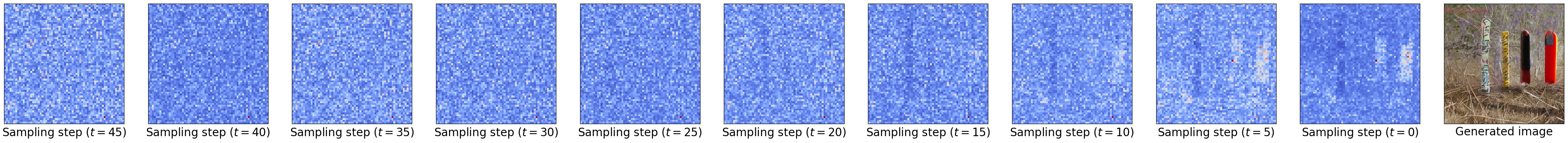}

  \caption{Uncertainty maps obtained from our proposed method. Coherently to our findings in the main article (Figure 3 in the main article), we observe high uncertainty in the first phases of the sampling process with very little differences between different samples, while most of the uncertainty related to the elements in the final image are in the last steps of the denoising process.}
\end{figure*}

\begin{figure*}
  \includegraphics[width=0.99\linewidth]{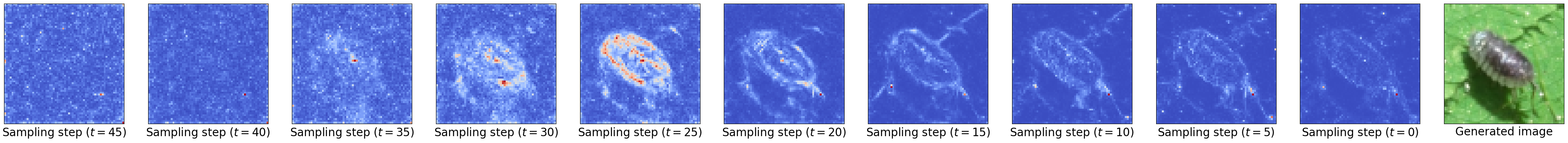}
  \includegraphics[width=0.99\linewidth]{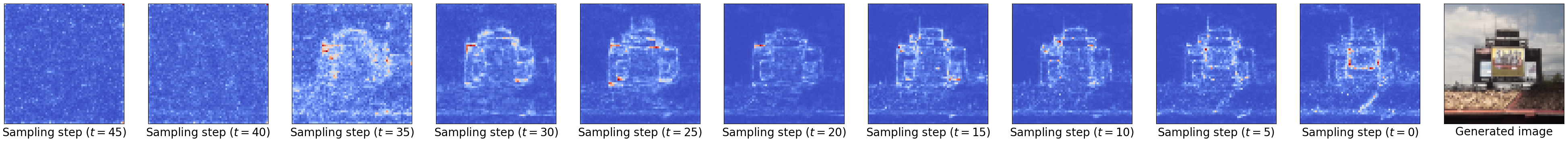}
  \includegraphics[width=0.99\linewidth]{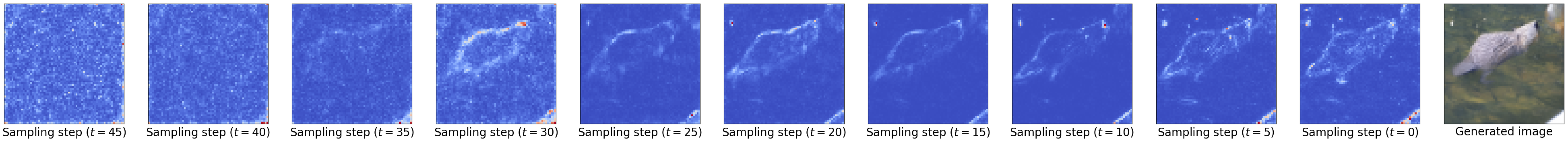}
  \label{fig:mc_dropout}
  \caption{Uncertainty maps obtained from MC Dropout. While our method has high uncertainty on foreground objects, we observe that MC-Dropout has high uncertainty only on the edges of foreground objects}
  
\end{figure*}


\begin{figure*}
  \includegraphics[width=0.99\linewidth]{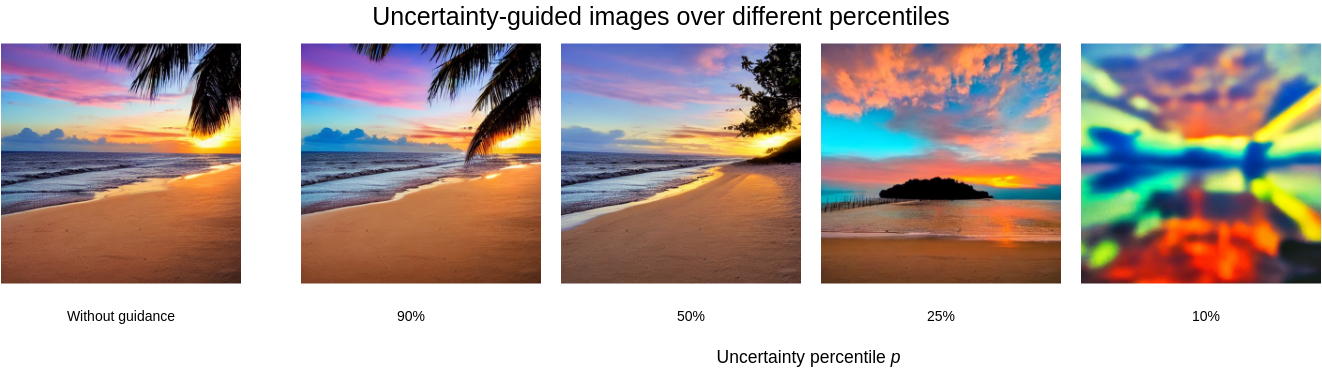}
  \vfill
  \vspace{2em}
  \includegraphics[width=0.99\linewidth]{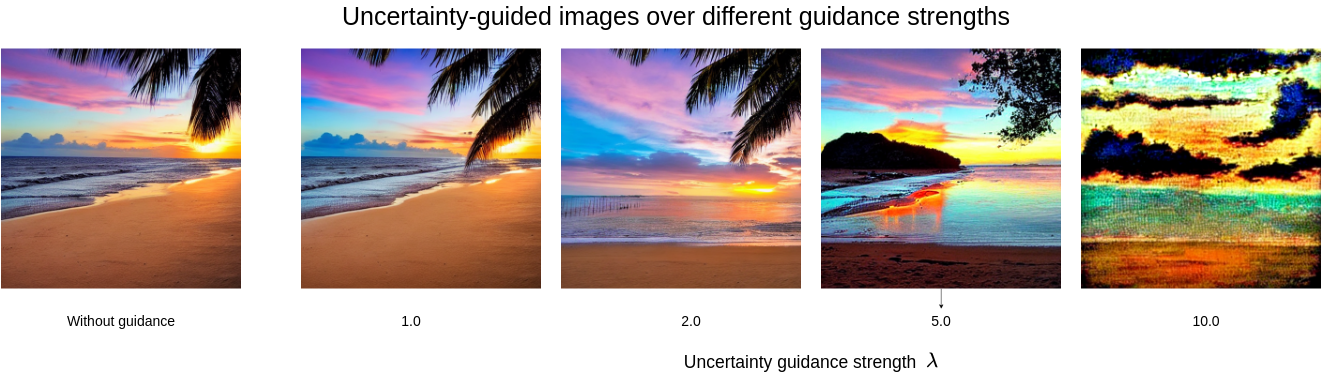}
  \caption{Hyperparameter sweep for uncertainty-guided sampling on Stable Diffusion 1.5. The top row shows the effect of varying the uncertainty percentile threshold, while the bottom row demonstrates the impact of adjusting the uncertainty strength. In the first row, by lowering the uncertainty percentile \textit{p} we change important scene details as the sun. In the second row, by increasing the uncertainty guidance strength $\lambda$, we are fundamentally changing the scene structure.}
  \label{fig:uncertainty_sweep}
\end{figure*}

\begin{figure*}
  \centering
  \includegraphics[width=0.3\linewidth]{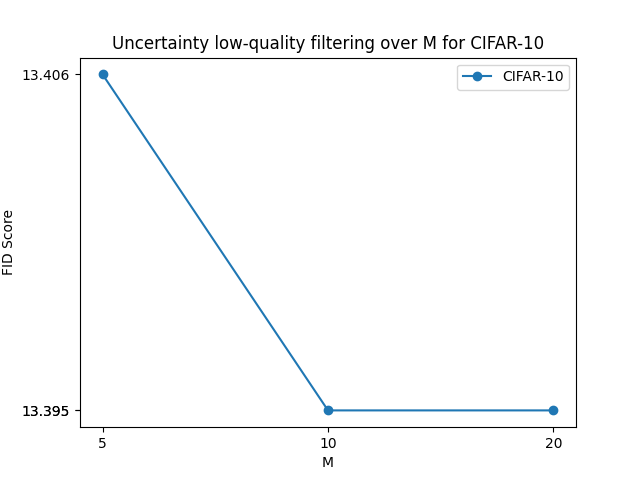}
  \includegraphics[width=0.3\linewidth]{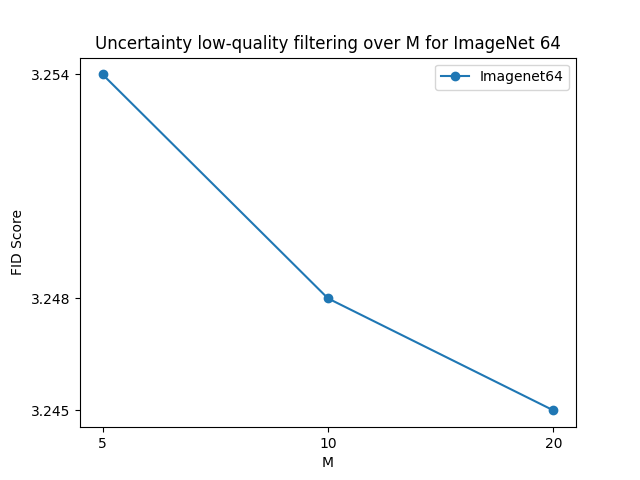}
  \includegraphics[width=0.3\linewidth]{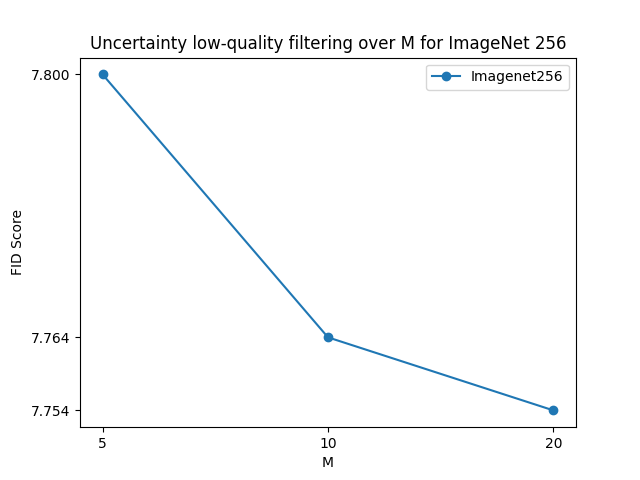}

  \label{fig:time_M}
  \caption{Uncertainty low quality filtering as in Table 1 of the main article, but using different number of perturbated samples for uncertainty estimation (M). We observed slight improvements with higher, but at the cost of higher prediction times as highlighted by Table \ref{tab:time-generation} and \ref{tab:time-generation_m_20}}
\end{figure*}

\begin{figure*}
  \centering
  \includegraphics[width=0.85\linewidth]{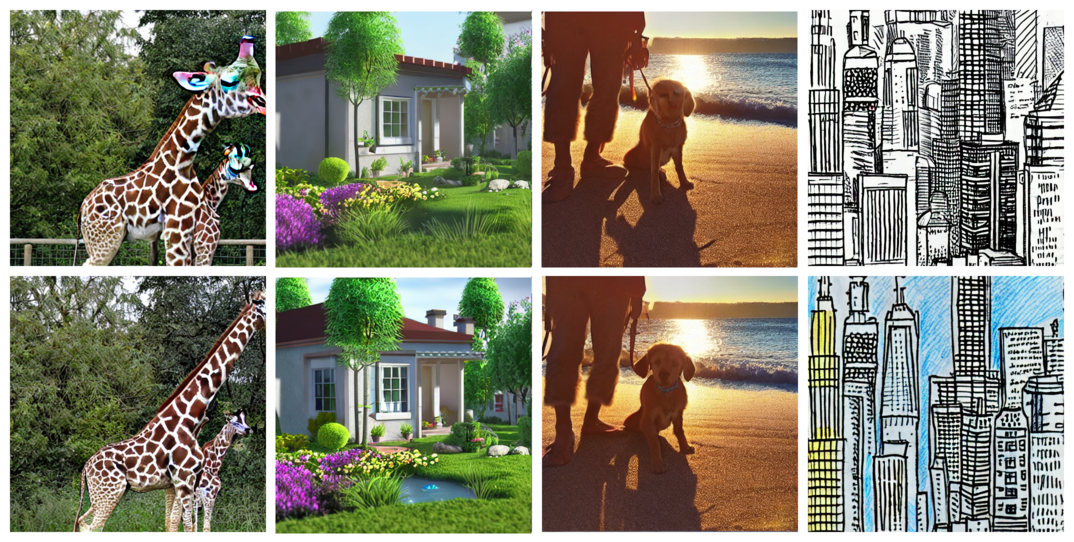}
  \label{fig:aditional_vis_res_3}
  \caption{Additional visual results of uncertainty guidance applied to Stable Diffusion. For each pair of images, the \textit{top row} shows the generated image \textit{without} uncertainty guidance while the \textbf{bottom row} shows the same image generated \textbf{with} uncertainty guidance.}
\end{figure*}

\begin{figure*}
  \centering
  \includegraphics[width=0.85\linewidth]{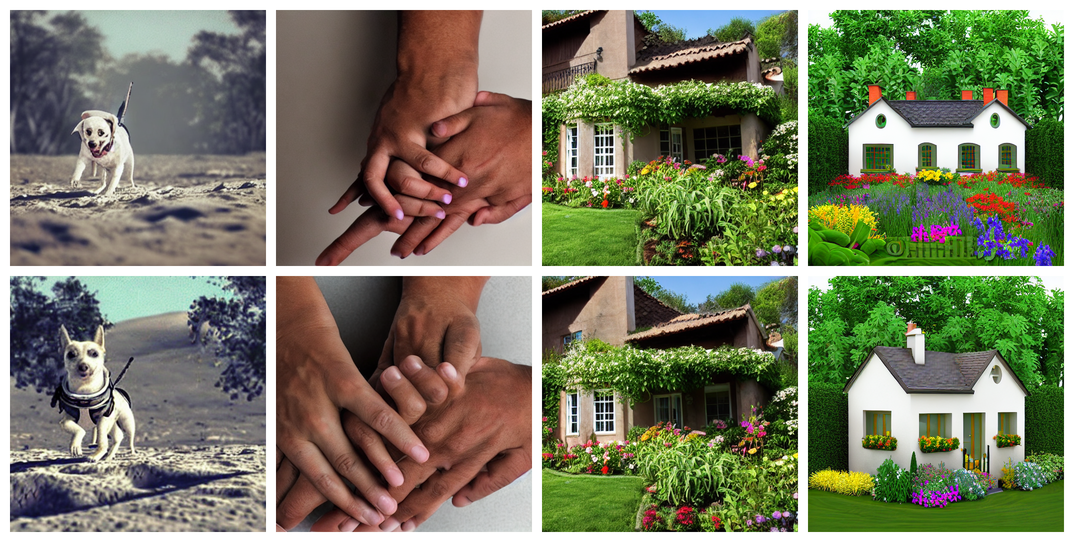}
  \label{fig:aditional_vis_res_4}
  \caption{\blue{Additional visual results of uncertainty guidance applied to Stable Diffusion. For each pair of images, the \textit{top row} shows the generated image \textit{without} uncertainty guidance while the \textbf{bottom row} shows the same image generated \textbf{with} uncertainty guidance. In the second column we observe the failure of the uncertainty guidance with human hands, as generating coherent hands is a very challenging task for Stable Diffusion. In the third column we observe very small changes with the uncertainty guidance, as the generated image is already of high quality. However, with hyper-parameter tuning, we can observe further improvements as demonstrated in Figure \ref{fig:uncertainty_sweep} }}
\end{figure*}

\section{Additional tables}

\begin{table*}[t]
  \caption{\blue{Comparison of the Precision and Recall between $ 60\,000 $ generated images with and without the uncertainty guidance, except for Imagenet512 for memory reasons. }}
  \vspace{-.1cm}
  \label{tab:filtering-results}
  \setlength{\tabcolsep}{0.55em} %
  \small
  \centering
  \begin{tabular}{ccrrrr}
  \toprule
  \multirow{2}{*}{Model}&\multirow{2}{*}{Dataset}& \multicolumn{2}{c}{\text{Precision} $\uparrow$} & \multicolumn{2}{c}{\text{Recall} $\uparrow$} \\
  \cmidrule{3-6} & & Random & Ours & Random & Ours  \\
  \midrule
  ADM & ImageNet 64 & 0.999 & 0.999 & 0.004 & 0.005 \\
  ADM & ImageNet 128 & 0.951 & 0.951 & 0.371 & \textbf{0.380} \\
  ADM w/2-DPM & ImageNet 128 & \textbf{0.874} & 0.872 & 0.524 & \textbf{0.540} \\
  U-ViT & ImageNet 256 & 0.325 & \textbf{0.339} & 0.762 & \textbf{0.856} \\
  U-ViT &ImageNet 512 & 0.791 & \textbf{0.793} & 0.431 & \textbf{0.451} \\
  DDPM & CIFAR-10 & 0.685 & 0.685 & 0.00 & 0.00 \\
  \bottomrule
  \end{tabular}
  \vspace{-.2cm}
\end{table*}

\begin{table*}[t]
  \caption{{Comparison of generation time with and without uncertainty estimation in seconds of 128 samples, using the same setup described in Section 4.1 of the main article, i.e. using M=5, 50 generation steps and compute the uncertainty between step 45 and 48.}}
  \vspace{-.1cm}
  \label{tab:time-generation}
  \setlength{\tabcolsep}{0.55em} %
  \small
  \centering
    \begin{tabular}{ccrr}
    \toprule
    \multirow{2}{*}{Model}&\multirow{2}{*}{Dataset} & \multicolumn{2}{c}{\text{M=5}} \\
     &  & Without uncertainty estimation & With uncertainty estimation \\
    \midrule
    ADM&ImageNet 64& 40.753 & 52.387  \\
    ADM&ImageNet 128& 86.805 & 112.777 \\
    ADM w/2-DPM &ImageNet 128& 86.712 & 112.765 \\
    U-ViT&ImageNet 256& 26.272 & 37.058  \\
    U-ViT&ImageNet 512& 32.859 & 47.531  \\
    DDPM & CIFAR-10 & 2.661 & 3.671 \\
    \bottomrule
  \end{tabular}
  \vspace{-.2cm}
\end{table*}

\begin{table*}[t]
  \caption{{Comparison of generation time with and without uncertainty estimation in seconds of 128 samples, using the same setup described in Section 4.1 of the main article, i.e. except for M=20, 50 generation steps and compute the uncertainty between step 45 and 48.}}
  \vspace{-.1cm}
  \label{tab:time-generation_m_20}
  \setlength{\tabcolsep}{0.55em} %
  \small
  \centering
    \begin{tabular}{ccrr}
    \toprule
    \multirow{2}{*}{Model}&\multirow{2}{*}{Dataset} & \multicolumn{2}{c}{\text{M=20}} \\
     &  & Without uncertainty estimation & With uncertainty estimation \\
    \midrule
    ADM&ImageNet 64& 41.013 & 89.316  \\
    ADM&ImageNet 128& 86.768 & 190.939 \\
    ADM w/2-DPM &ImageNet 128& 86.750 & 190.871 \\
    U-ViT&ImageNet 256& 43.987 & 60.550  \\
    U-ViT&ImageNet 512& 53.189 & 74.420  \\
    DDPM & CIFAR-10 & 2.726 & 6.302 \\
    \bottomrule
  \end{tabular}
  \vspace{-.2cm}
\end{table*}

\end{document}